\documentclass{IEEEtran}

\IEEEoverridecommandlockouts
\usepackage{cite}
\usepackage{amsmath,amssymb,amsfonts}
\usepackage{algorithmic} 
\usepackage{graphicx} 
\usepackage{textcomp}
\usepackage{xcolor} 
\usepackage{subcaption}
\usepackage{hyperref}
\usepackage{verbatim}
\usepackage{lineno}
\def\BibTeX{{\rm B\kern-.05em{\sc i\kern-.025em b}\kern-.08em
    T\kern-.1667em\lower.7ex\hbox{E}\kern-.125emX}}
\begin{document}

\title{Learning Alzheimer’s Disease Signatures by bridging EEG with Spiking Neural Networks and Biophysical Simulations
}

\author{\IEEEauthorblockN{Szymon Mamoń}

\IEEEauthorblockA{\textit{Department of Computer Science} \\
\textit{AGH University of Krakow}, 
Krakow, Poland 
}
\and

\IEEEauthorblockN{  Max Talanov}

\IEEEauthorblockA{\textit{Department of Mathematics and Informatics} \\
\textit{University of Messina}, 
Messina, Italy. 
} 

\and
\IEEEauthorblockN{  Alessandro Crimi}

\IEEEauthorblockA{\textit{Department of Computer Science} \\
\textit{AGH University of Krakow}, 
Krakow, Poland 
}
 
}

\maketitle
\begin{abstract}
As the prevalence of Alzheimer's disease (AD) continues to increase, improving mechanistic understanding from non-invasive biomarkers is increasingly critical. Recent work suggests that brain circuit-level alterations manifest as measurable changes in electroencephalography (EEG) spectral features that can be detected by machine learning models. However, conventional deep learning approaches for EEG-based AD detection are computationally intensive and mechanistically opaque. Spiking neural networks (SNNs) offer a biologically plausible and energy-efficient alternative, yet their application to AD diagnosis remains largely unexplored. 
We propose a neuro-bridge framework that explicitly links data-driven learning with minimal yet biophysically grounded simulations, enabling bidirectional interpretation between machine learning signatures and circuit-level mechanisms in AD. Using clinical resting-state EEG data, we train an SNN classifier that  achieves competitive accuracy (AUC=0.839), while highlighting aperiodic 1/f slope as one of the key discriminative markers.
 
The 1/f slope is a macroscopic proxy of the excitation–inhibition balance. To mechanistically interpret these findings, we construct spiking network simulations in which inhibitory-to-excitatory synaptic ratios are systematically varied to emulate healthy, mild cognitive impairment, and AD-like states. By comparing membrane potential-based and synaptic current-based EEG proxies, we show that both recapitulate empirical spectral slowing and altered alpha organization. 
Incorporating empirical functional connectivity priors into multi-subnetwork simulations enhances spectral differentiation, demonstrating that large-scale network topology constrains EEG signatures more strongly than excitation–inhibition balance alone. 
Overall, this "neuro-bridge" approach connects SNN-based classification with interpretable circuit simulations, advancing the mechanistic understanding of EEG biomarkers while enabling scalable, explainable AD detection.
\end{abstract} 


\section{Introduction}
Despite decades of research, early detection of Alzheimer's disease (AD) in routine clinical settings remains challenging, as current diagnostic approaches often rely on expensive neuroimaging techniques or invasive biomarker assays which are not easily scalable \cite{nichols2022estimation}.  Electroencephalography (EEG) is a non-invasive technique with relatively low acquisition and operational costs, and its value for the diagnosis and monitoring of AD has been well established \cite{dauwels2010diagnosis}. However, the development of interpretable, computationally efficient, and mechanistically grounded EEG-based diagnostic tools still remains an open challenge \cite{akbar2025unlocking}.

It has been shown that excitation–inhibition (E/I) balance is a fundamental organizational principle of cortical circuits that can be quantified using macroscopic electrophysiological signals such as EEG and its inverse power slope \cite{van2023resting}. The E/I balance governs the operating regime of cortical networks, with healthy circuits maintaining near-critical dynamics that maximize information processing capacity while preventing pathological states such as runaway excitation or excessive synchrony. Disruptions in E/I balance have been implicated in numerous neurological and psychiatric disorders and neurodegeneration, where progressive loss of synaptic functions and network dysfunction lead to observable changes in oscillatory dynamics \cite{palop2016network}.

Recent works introduced and validated a functional EEG-derived measure of excitation–inhibition balance  using computational models of critical oscillatory networks \cite{diachenko2024functional}.
 These findings strongly motivate approaches that combine spiking network models with EEG signal analysis: if network simulations can recapitulate features captured by E/I balance and related spectral measures, such models may serve as mechanistic bridges between microscopic E/I imbalance and macroscale biomarkers observed in clinical conditions such as AD.

Parallel to these advances in understanding the E/I balance, spiking neural networks (SNNs) have emerged as a promising approach for temporal signal processing. SNNs encode information on the precise timing of discrete events (spikes) rather than continuous activations, offering biological plausibility, strong temporal modeling capabilities, and markedly lower energy consumption on neuromorphic hardware \cite{mikhaylov2020neurohybrid}. These properties make SNNs especially suitable for EEG-based tasks, where temporal structure, computational efficiency, and real-time processing constraints are paramount \cite{khan2025review}.

The application of SNNs to EEG analysis has gained traction in multiple domains. Yuan et al. \cite{CAI2026108127} introduced a wavelet-enhanced spiking transformer for emotion recognition and auditory attention decoding , demonstrating that multi-scale time–frequency representations can be integrated directly into spiking models without handcrafted feature engineering. In the clinical domain, Burelo et al. developed an SNNs-based detector for high-frequency oscillations in epilepsy \cite{Burelo2022}, showing that SNN can identify subtle EEG biomarkers with strong clinical relevance. Other studies \cite{Zhang2024} further demonstrated that recurrent spiking architectures achieve cross-patient seizure detection with performance comparable to traditional artificial neural networks (ANNs) while reducing theoretical energy consumption by several orders of magnitude. Most importantly, a real-time seizure detection on the Xylo neuromorphic chip has been implemented \cite{LI2024109225} , achieving sub-milliwatt power consumption while maintaining high accuracy, illustrating the feasibility of practical neuromorphic EEG monitoring systems.

Despite these successes in epilepsy research and related applications, the potential of SNNs for the detection of Alzheimer's disease remains largely unexplored. 
Even though AD and epilepsy are clinically intertwined \cite{dun2022bi}.  This gap is surprising given that AD presents many of the same challenges that SNNs have successfully addressed in other domains: non-stationary signals, subtle temporal features embedded in oscillatory dynamics, and the need for energy-efficient, deployable diagnostic tools. In addition, existing deep learning approaches to AD classification using EEG typically operate as black boxes, offering limited information on the neurobiological mechanisms underlying disease-related spectral changes.

The present work addresses these gaps by developing a unified framework that integrates SNN-based EEG classification with biophysically grounded network simulations of E/I imbalance. We call this a \textit{"neuro-bridge"}.  
Our approach is motivated by three key observations. First, aperiodic spectral features, particularly the slope of 1/f, have been associated with cortical E/I balance and show alterations in AD \cite{Voytek2015}. Second, an SNN classifier provides a friendly substrate for modeling E/I dynamics because of their explicit representation of excitatory and inhibitory populations. Third, mechanistic simulations can complement data-driven classification by revealing how microscale circuit alterations propagate to macroscopic EEG signatures. Linking EEG classification to power spectral features and the excitatory–inhibitory ratio allows us to move from black-box prediction to mechanistically interpretable biomarkers grounded in cortical circuit dynamics.

\begin{figure}[htbp]
    \centering
    \includegraphics[width=\columnwidth]{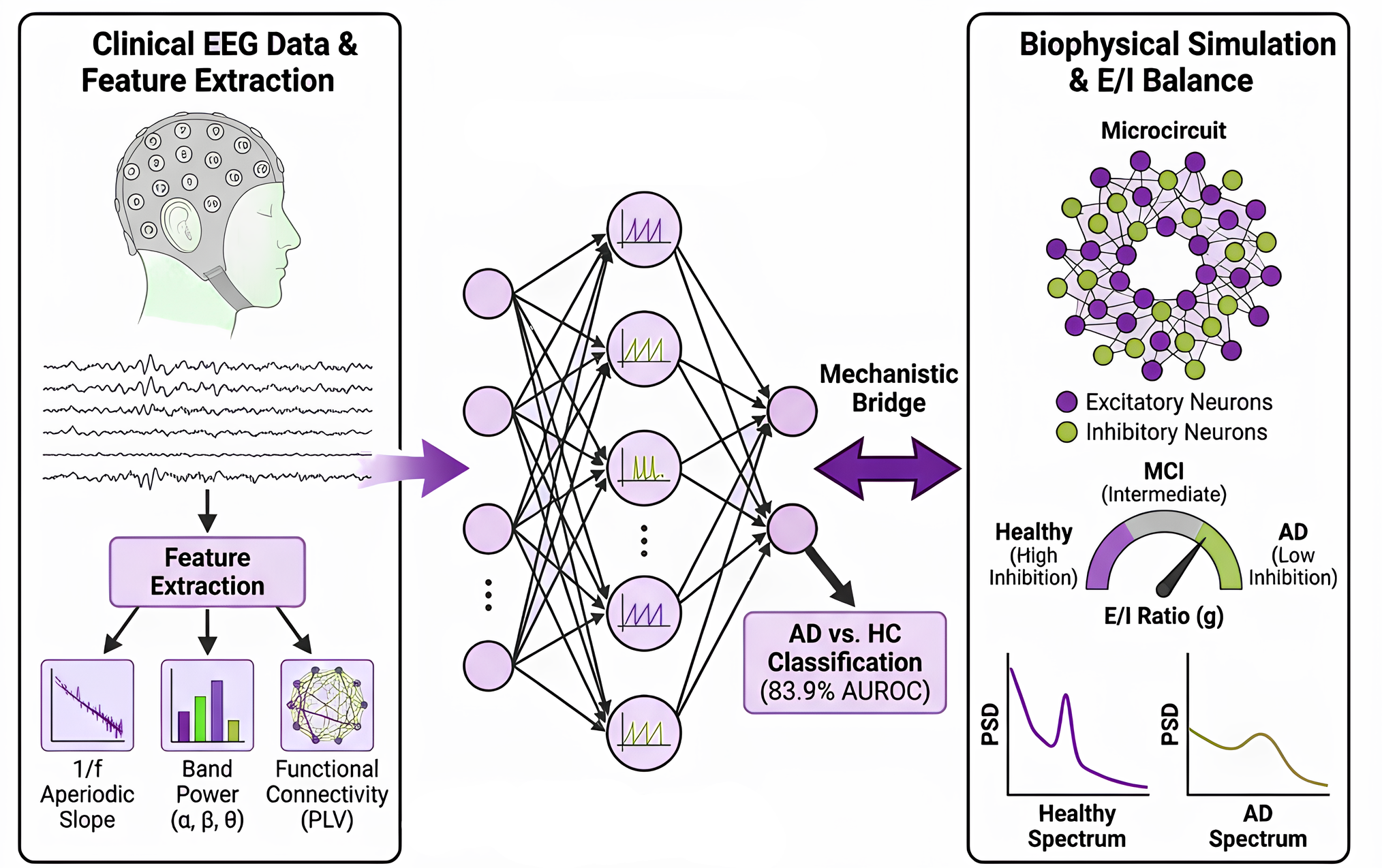}
    \caption{Simplistic overview of the connections between SNN classifier and minimal biophisical simulation.}
    \label{fig:pipeline}
\end{figure}

We propose a spiking neural network framework that jointly models and classifies Alzheimer's disease EEG signatures, linking learning-based SNN classification with mechanistic E/I balance simulations as depicted in Figure \ref{fig:pipeline}. Therefore, the purpose of this paper is i) to compare the validity of the inverse power slope as a signature to identify neurodegeneration with neuromorphic AI architectures, ii) to relate this signature with simple mechanistic simulations based on the ratio between inhibitory and excitatory neurons with realistic simulations,
and iii) to investigate other simulations that also include functional connectivity priors  from the same EEG dataset to investigate whether functional connectivity is a prior stronger than the ratio between inhibitory and excitatory neurons. 

\section{Methods}

\subsection{Data and pre-processing}
In the reported experiments, we used human EEG data and computationally generated data by the Nest simulator \footnote{https://www.nest-simulator.org/}, and we invite the reader to the simulations subsections for the details. 

Human data comes from a publicly available dataset \cite{miltiadous2023dataset} comprising routine clinical scalp EEG recordings. This includes 36 patients diagnosed with Alzheimer’s disease and 29 healthy controls of cognitively normal age-matched. 
EEG data were acquired during resting-state, eyes-closed conditions. The recordings were performed using a Nihon Kohden clinical EEG system with 19 scalp electrodes placed according to the international 10–20 system, along with mastoid reference electrodes (A1–A2), with durations ranging from approximately 5 to 20 minutes. Data provided in BIDS (Brain Imaging Data Structure) format were  pre-processed using the Butterworth band-pass filter (0.5–45 Hz), and after the signals were re-referenced to A1-A2. Then, artifact subspace reconstruction (ASR) was used to remove segments exceeding a 0.5 s window with a standard deviation above 17, considered a conservative threshold.
Following ASR, independent component analysis (ICA) was applied for the removal of EEG artifacts. We use an approximate source-level inference \cite{rojas2018study} where sensor-based connectivity is converted into source-based using sensors as seeds.
The code used to perform the analysis described in the following
sub-sections is publicly available at the URL \href{https://github.com/alecrimi/neurobridge}{https://github.com/alecrimi/neurobridge} . 

\subsection{EEG Classification}
EEG signals were classified using a three-layer SNN architecture implemented in \texttt{snntorch}\footnote{https://github.com/snntorch}, with leaky integrate-and-fire neurons and sigmoidal surrogate gradients. Rate-based encoding was applied to convert continuous EEG features into spike trains, creating temporal event-driven representations of neural activity. To benchmark the SNN, a corresponding artificial neural network (ANN) with a similar structure 3-layer dense multi-layer perceptron was trained on the same features, allowing direct comparison of the two approaches.
Our primary goal is mechanistic interpretation rather than maximum classification accuracy. This objective requires engineered features that are interpretable and biologically meaningful, enabling comparable analyzes in empirical EEG and simulated network data. Feature engineering also provides practical advantages: standard ANNs struggle with raw time series without recurrent architectures or transformer models, while SNNs, though better suited for temporal data, require very large datasets for end-to-end learning from raw signals.
Our pipeline therefore proceeds as follows: EEG recordings are segmented into epochs, from which we extract neurophysiologically relevant features. These features are then fed directly to the ANN, or first encoded into spike trains before being processed by the SNN. Specifically, the input feature set included spectral power bands, standard deviation (signal amplitude variability),  functional connectivity metrics, and aperiodic components. Among these, the 1/f slope, computed using the Fitting Oscillations and One-Over-F (FOOOF) algorithm \cite{donoghue2020parameterizing}, was highlighted using SHapley Additive exPlanations (SHAP) analysis to quantify its relative contribution to model predictions \cite{lundberg2017unified}. In the SHAP feature importance rankings, node-specific features are marked using the same electrode sensor labels as in the EEG montage. Even though the The 1/f slope  might not be the most relevant feature, for our purpose is critical as the The 1/f slope is a macroscopic proxy of the excitation–inhibition (E/I) balance which is investigated later by the simulations.
 
\begin{figure}[htbp]
    \centering
    \includegraphics[width=\columnwidth]{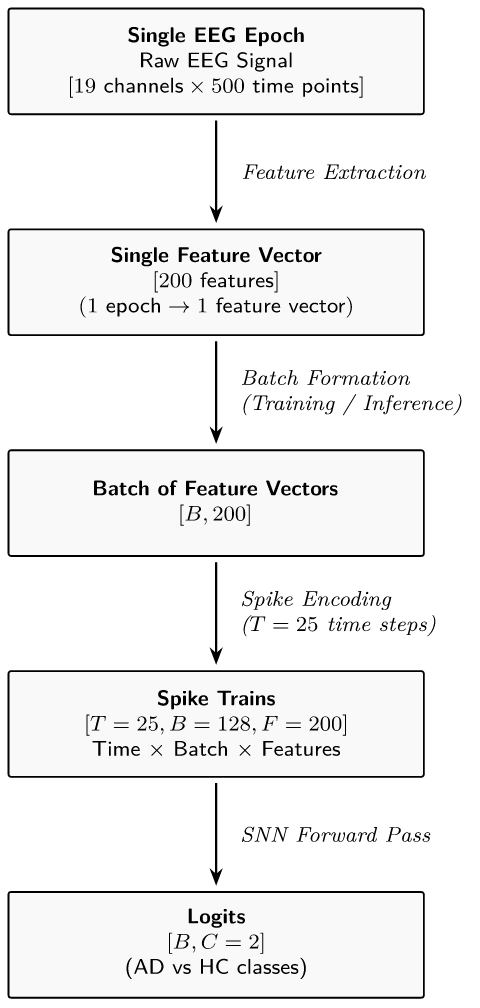}
    \caption{SNN-based EEG classification pipeline. Raw EEG epochs (19 channels × 500 samples), with each epoch/window having 500 time points. Those then undergo feature extraction to compute spectral power bands, phase-locking values (PLV), and aperiodic slope (1/f). The resulting 200-dimensional feature vectors are batched (B=128), encoded into spike trains via rate-based encoding over T=25 time steps, and classified by a 3-layer spiking neural network with leaky integrate-and-fire neurons. The network outputs class predictions C (AD vs. HC) for each epoch in the batch. For the ANN classifier the pipeline is the same except the spike train encoding.}
    \label{fig:snn_workflow}
\end{figure}

Classification was performed at window-level (also called epochs), using individual EEG segments as input as depicted in Figure \ref{fig:snn_workflow}. This approach preserves the temporal granularity inherent in spiking representations and facilitates the interpretation of the use of the SNN feature. Overall, this framework demonstrates how SNNs can integrate multi-feature EEG data for neurodegenerative classification while providing mechanistic insights, and a comparison with traditional ANN architectures. However, the purpose of this classifier is not to outperform a corresponding ANN, but to examine how SNNs encode EEG signals and reveal mechanistic insights related to E/I balance. The ANN comparison provides context, while feature analyzes highlight biologically significant contributions within the spiking framework.

\subsection{Network based statistics for EEG sub-bands}
From the EEG data, functional connectivity is defined by phase-lock value (PLV) estimators for each subject. Given those functional networks, we compare the two populations of connectivity using the network-based statistics (NBS) to further investigate the statistically significant different connections. NBS is a nonparametric permutation-based statistical framework designed to identify connected sub-nets of brain connectivity that differ significantly between groups while controlling the family error rate \cite{Zalesky2010NBS}. 
Here, we performed a band-specific NBS analysis to identify statistically significant differences in brain connectivity networks between the two groups of healthy controls (HC) and   AD. This step is performed first as a preliminary analysis which should match what we see in the classification and simulation. Later, the  functional connectivity computed in this step will also be used as a prior for a more refined simulation.

More precisely, the NBS analysis is repeated independently for multiple EEG frequency bands:
$
\mathcal{B} = \{ \delta, \theta, \alpha, \beta, \gamma\}$, 

The $\delta$ band is sometimes avoided to define connectivity between regions using an EEG signal. Indeed, there is evidence that PLV estimators are biased and have a larger variance in the low frequency bands and can give a noisy estimator \cite{vinck2010pairwise}.  We have included the evaluation for this band for consistency with the simulation experiments in literature (e.g. \cite{MartinezCanada2023})  which instead consider it generally, as we want to make a link between those fields, though we recommend care in conclusions for this band.


For each subject, the functional connectivity is computed for all four $\mathcal{B}$  bands.  More specifically, the functional connectivity is estimated using the PLV, which measures the temporal stability of phase differences between pairs of EEG signals \cite{blanco2024investigating}.  
Briefly, given two band-pass filtered signals $X(t)$ and $Y(t)$, their instantaneous phases $\phi_X(t)$ and $\phi_Y(t)$ were extracted using the Hilbert transform. 
PLV is defined as 
\begin{equation}
\mathrm{PLV}_{XY} = \left| \frac{1}{T} \sum_{t=1}^{T} 
\exp\left( i \left[ \phi_X(t) - \phi_Y(t) \right] \right) \right|, 
\end{equation}
where $T$ denotes the number of time samples and $i$ is the imaginary unit. 
PLV values range from 0, indicating no consistent phase relationship, to 1, indicating perfect phase synchronization. In this way, we have connectivity matrices  $\mathbf{C}^{(s)} \in \mathbb{R}^{N \times N}$, where
 $N$ is the number of channels (nodes), and $s$ indexes subjects.  If we further consider  each group $g \in \{\mathrm{HC}, \mathrm{AD}\}$ and band $b \in \mathcal{B}$, we can consider:
\[
\mathcal{C}_{g,b} =
\left\{
\mathbf{C}^{(1)}, \mathbf{C}^{(2)}, \dots, \mathbf{C}^{(n_g)}
\right\},
\]
where $n_g$ is the number of subjects in the group $g$.

For each frequency band, we perform two independent NBS tests corresponding to the two possible one-sided hypotheses, since for each band connectivity might be decreased or increased for the AD patients:

\subsubsection{Left-Tailed Test}
\[
H_0: \mu_{\mathrm{HC}} \leq \mu_{\mathrm{AD}}, \quad
H_1: \mu_{\mathrm{HC}} > \mu_{\mathrm{AD}},
\]
testing for decreased connectivity in AD.

\subsubsection{Right-Tailed Test}
\[
H_0: \mu_{\mathrm{HC}} \geq \mu_{\mathrm{AD}}, \quad
H_1: \mu_{\mathrm{AD}} > \mu_{\mathrm{HC}},
\]
testing for increased connectivity in AD.
 
For each edge $(i,j)$, a test statistic (typically a two-sample $t$-statistic) is computed:
\begin{equation}
    T_{ij} =
\frac{\bar{C}^{\mathrm{HC}}_{ij} - \bar{C}^{\mathrm{AD}}_{ij}}
{\sqrt{\frac{\sigma^2_{\mathrm{HC},ij}}{n_{\mathrm{HC}}}
+ \frac{\sigma^2_{\mathrm{AD},ij}}{n_{\mathrm{AD}}}}},
\end{equation}

Edges satisfying:
\[
|T_{ij}| > T_{\text{primary}},
\quad T_{\text{primary}} = 3,
\]
are retained, forming a thresholded graph. The suprathreshold graph is decomposed into connected components:
\[
\mathcal{G}_1, \mathcal{G}_2, \dots, \mathcal{G}_K,
\]
where each component represents a candidate subnetwork.
 
For each observed component $\mathcal{G}_k$, a family-wise error corrected $p$-value is computed, and is considered significant if $p_k < \alpha, \quad \alpha = 0.01$. The results are stored separately for each $(\text{band}, \text{tail}) \in \mathcal{B} \times \{\text{left}, \text{right}\}$, and are reported in Table \ref{tab:nbs_edges}, the statistically significant connections are reported in Figure \ref{fig:NBS}.  Apart being a further validation of the different signals between HC and AD, those connectivity features will be used in the latter experiments for the connectivity-informed simulations. 

\subsection{Spiking Network models for simulating E/I imbalance effects on spectral properties} 
Following EEG classification based on power spectral features, the E/I ratio was estimated by fitting a simplistic yet biologically grounded spiking neural network model to the observed spectral parameters and quantifying the relative contributions of excitatory and inhibitory synaptic activity required to reproduce empirical EEG spectra. 

We used randomly connected networks to study excitatory/inhibitory imbalance rather than full laminar cortical models. Random networks capture the essential statistical and dynamical properties of cortical populations including irregular asynchronous firing, oscillatory dynamics, and spectral scaling, while maintaining computational tractability and interpretability. 
We used 400 neurons as the minimum reservoir size required for the described experiments to reproduce the observed effects. We considered several model configurations, including one in which the 400 neurons are distributed across 19 areas corresponding to the 19 EEG channels (see Subsection \ref{sec:bands}), interpreted as at least 19 cortical columns. In a simplified cortical column model, each column was assumed to contain three neuronal groups corresponding to cortical Layers 2-3, Layer 4, and Layers 5-6. Each group was further assumed to include both excitatory and inhibitory neurons, with a $80\times20\%$ excitatory-to-inhibitory ratio. To maintain excitation–inhibition balance in healthy controls, we required at least one inhibitory neuron per layer. This constraint results in a minimum of six excitatory neurons per neuronal group. With three groups per column and 19 cortical columns, this yields approximately 400 neurons in total. 
While this represents a highly simplified abstraction of cortical organization, it provides a plausible lower bound on the number of randomly connected neurons required to reproduce neurodegenerative effects associated with AD. Moreover, few hundreds neurons are commonly used in spiking-network simulations for the same purpose \cite{vogels2005signal,MartinezCanada2023}. 
 
The randomly connected network approach is widely used in the literature to investigate the effects of E/I perturbations on population-level signals and EEG-like activity \cite{Brunel2000,vogels2005signal,MartinezCanada2023}. 
Using more detailed laminar models would increase the complexity of the parameters without necessarily improving the correspondence with the biomarkers of interest in EEG \cite{gao2017inferring}. This relevant aspect is nevertheless further debated in the Discussion section. 

We considered two types of network-derived signals: a membrane-potential–based network signal and a synaptic-current–based network signal. Both models systematically varied the inhibitory-to-excitatory synaptic strength ratio $g = g_I/g_E$ across conditions representing AD (low inhibition), mild cognitive impairment (MCI, intermediate inhibition), and HC (high inhibition).
 Based on the literature \cite{Brunel2000,MartinezCanada2023},  we defined  specific ratios as reported in Table \ref{tab:ei_mapping}, and performed a simulation using the Nest tool \cite{Eppler2008PyNEST}.  The MCI simulations are included for consistency with the literature, although they cannot be directly compared with the EEG data in our data set because there are no EEG data for MCI patients. The simulations are run 10 times with different random connections, and the averaged value is computed. 

\begin{table}[!t]
\centering
\caption{Mapping between excitation--inhibition (E/I) balance and spiking network regimes. The inhibitory-to-excitatory synaptic strength ratio $g = g_I/g_E$ is varied following canonical balanced network models.}
\label{tab:ei_mapping}
\begin{tabular}{llll}
\hline
\textbf{Condition} & \textbf{$g = g_I/g_E$} & \textbf{Network Regime} & \textbf{EEG Signature} \\
\hline
AD  & 2.5 & Excitation-dominance & Flatter 1/f \\
MCI & 3.5 & Near-critical   & Intermediate \\
HC  & 6.5 & Inhibition-dominance & Steeper 1/f \\
\hline
\end{tabular}
\end{table}

\subsubsection{Model 1: Membrane-potential–based network signal}
The first model extracts a network-level signal derived from simulated neuronal membrane potentials. We implemented a network of 400 conductance-based leaky integrate-and-fire neurons (80\% excitatory, 20\% inhibitory) with heterogeneous Poisson-driven synaptic input and sparse recurrent connectivity (connection probability $p = 0.2$) \cite{MartinezCanada2023}. The balance of excitation and inhibition was manipulated by adjusting inhibitory synaptic strength $g$, while excitatory synaptic strength was fixed at $g_E = 2.0$ nS, following \cite{MartinezCanada2023}.  

Following \cite{Brunel2000}, neuronal heterogeneity in membrane potentials and firing thresholds was incorporated, together with structured external drive and background noise. After a 1000 ms warm-up period, membrane potentials from 20\% of randomly selected excitatory neurons were recorded at 1 kHz. A network signal was obtained by averaging these membrane potentials across neurons, and power spectral densities in the 0.5–40 Hz range were computed using Welch’s method with 4096-point segments and 75\% overlap. Spectra were normalized to relative power to emphasize changes in spectral shape.

\subsubsection{Model 2: Synaptic-current–based network signal}
The second model followed a complementary approach by simulating EEG proxies derived from synaptic currents, similar to \cite{MartinezCanada2023}. This network also comprised 400 integrate-and-fire neurons based on conductance (80\% excitatory, 20\% inhibitory) with exponentially decaying conductance-based synapses. Synaptic strengths (baseline $g_E = 2.0$ nS) and connection probabilities ($p = 0.2$) were also defined as in \cite{MartinezCanada2023} to maintain network stability across E/I regimes. Excitatory and inhibitory synaptic conductances ($g_{\text{ex}}$ and $g_{\text{in}}$) were recorded from 15\% of excitatory neurons  at 1 ms intervals. Synaptic currents were computed from the recorded conductances as $I_{\text{syn,ex}} = g_{\text{ex}}(V_m - E_{\text{ex}})$ and $I_{\text{syn,in}} = g_{\text{in}}(V_m - E_{\text{in}})$, where $E_{\text{ex}} = 0$ mV and $E_{\text{in}} = -80$ mV are the excitatory and inhibitory reversal potentials. An EEG proxy was computed as the population-averaged net synaptic current: $I_{\text{EEG}} = \langle I_{\text{syn,ex}} \rangle - \langle I_{\text{syn,in}} \rangle$, which approximates the extracellular potential measurable by scalp EEG. Power spectra (1–40 Hz) were computed using Welch's method with extended window sizes (8192 samples per segment, 93.75\% overlap). 
Spectra were normalized to their relative power.

The two models provide complementary perspectives on how E/I imbalance affects macroscopic neural signals. Model~1 emphasizes contributions from local membrane potential fluctuations. Model 2 focuses on synaptic current dipoles, which  which can be seen as a proxy of scalp-measured EEG signals. Both models shared core network architecture parameters (E/I ratio of 80:20, conductance-based neuron models, $g_E$ baseline) but differed substantially in  signal generation mechanisms, and spectral estimation methods, allowing us to assess the robustness of E/I effects across different biophysical signal origins and network configurations. All simulations were performed with condition-specific random seeds to ensure statistical independence across parameter sweeps. 

\subsection{Multi-Subnetwork Simulation with Band-Specific Functional Connectivity} \label{sec:bands}
Afterwards, we   implemented a large-scale spiking neural network simulation. The central idea is to run separate simulations for different frequency bands (theta, alpha, beta, gamma),  each constrained by a band-specific functional connectivity (FC) matrix, and then stitch together the resulting power spectra into a single composite spectrum. 
Even though structural connectivity would be the most straightforward basis to connect the sub-networks, 
Alzheimer's disease primarily affects functional connectivity even when structural connectivity might be relatively preserved in early stages. Functional disconnection often precedes structural degeneration, and it is therefore reasonable to use it \cite{al2024disrupted}. 
More specifically, each of the $N$ channels (nodes) represents a subnetwork with simular characteristics as described in the previous sections (leaky integrate-and-fire neurons with conductance-based synapses, and neurons divided into excitatory (E) and inhibitory (I) populations), and the level connectivity among the nodes for each class HC/AD and band gives the connection among the subnetworks.
More specifically, inter-subnetwork connections are driven by band-specific FC  normalized to $[0,1]$, 
yet coupling strength scales with FC magnitude,  
and connections are excitatory and target excitatory neurons only.  Thus, each simulation embeds empirical functional structure for one frequency band.  To maintain a simular number of total neurons, considering we are using 19 channels/subnetworks we  set each subnetwork having 21 neurons (achieving a total of 399 neurons). 

\subsection{Comparison of simulations with empirical EEG via Cohen's $d$}
It is possible to validate the simulations by comparing the changes in the subbands to the diffrences highlighted by NBS
for representing functional connectivity in the EEG data. Beyond this qualitative comparison, a quantitative validation can be given by comparing the sign and magnitude in difference for the  aperiodic exponent (1/f slope) between AD and HC, among simulations and empirical EEG signal, considered the ground-truth. 
More specifically, we computed Cohen's $d$ for each model and for the empirical EEG data, focusing on the aperiodic exponent (1/f slope), with the Cohen's $d$ defined as:

\begin{equation}
d = \frac{\mu_\text{AD} - \mu_\text{HC}}{s_\text{pooled}},
\end{equation}

where $s_\text{pooled}$ is the pooled standard deviation of the AD and HC groups:

\begin{equation}
s_\text{pooled} = \sqrt{\frac{(n_\text{AD}-1)\sigma_\text{AD}^2 + (n_\text{HC}-1)\sigma_\text{HC}^2}{n_\text{AD} + n_\text{HC} - 2}}.
\end{equation}

For the EEG analysis we used the empirical group means and standard deviations given by the two populations. For the simulations, we recreated this subject-level data, by running the simulations 30 times with different random connections.  

After computing Cohen's $d$ for each condition, we compared the sign and magnitude of the effect sizes across all simulations relative to the empirical EEG 1/f slope. This procedure allowed us to evaluate whether each simulation captures the direction of Alzheimer’s-related changes (i.e., whether the aperiodic exponent decreases in AD compared to HC) and to quantify how closely the simulated effect magnitudes align with the observed EEG effect sizes.

\section{Results}
In this section, we report first the experiments related to the classification and identification of discriminant functional connectivity features, and then the results related to simulations including those  connectivity-informed by the EEG signal. 

\subsection{EEG Classification}
\begin{figure}[!]
  \centering
  \includegraphics[width= \linewidth]{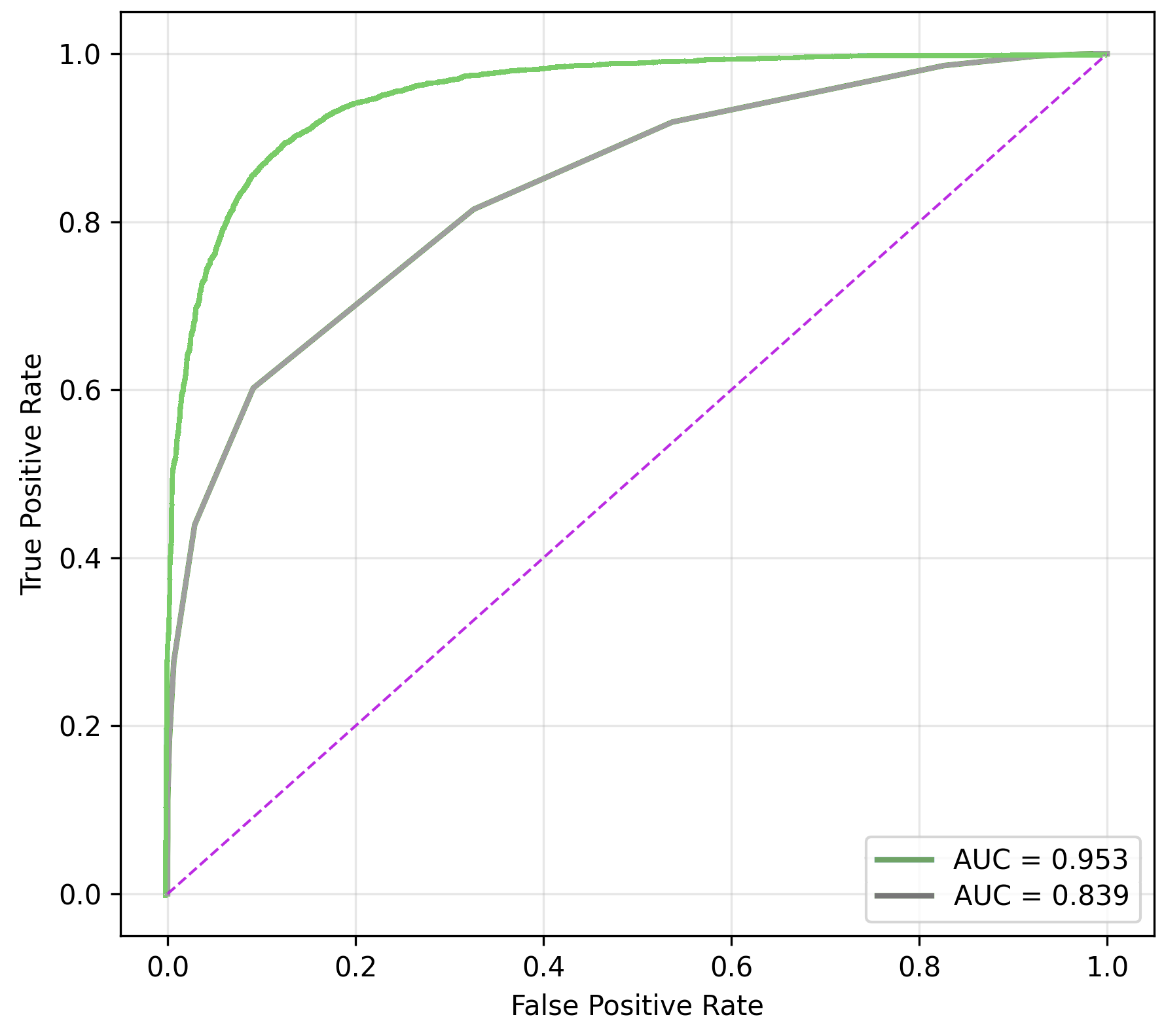} 
  \caption{ROC for both ANN (green line) and SNN (gray line) classifier}
  \label{AUC}
\end{figure}

\begin{figure*}[!]
  \centering
  \includegraphics[width=0.99 \linewidth]{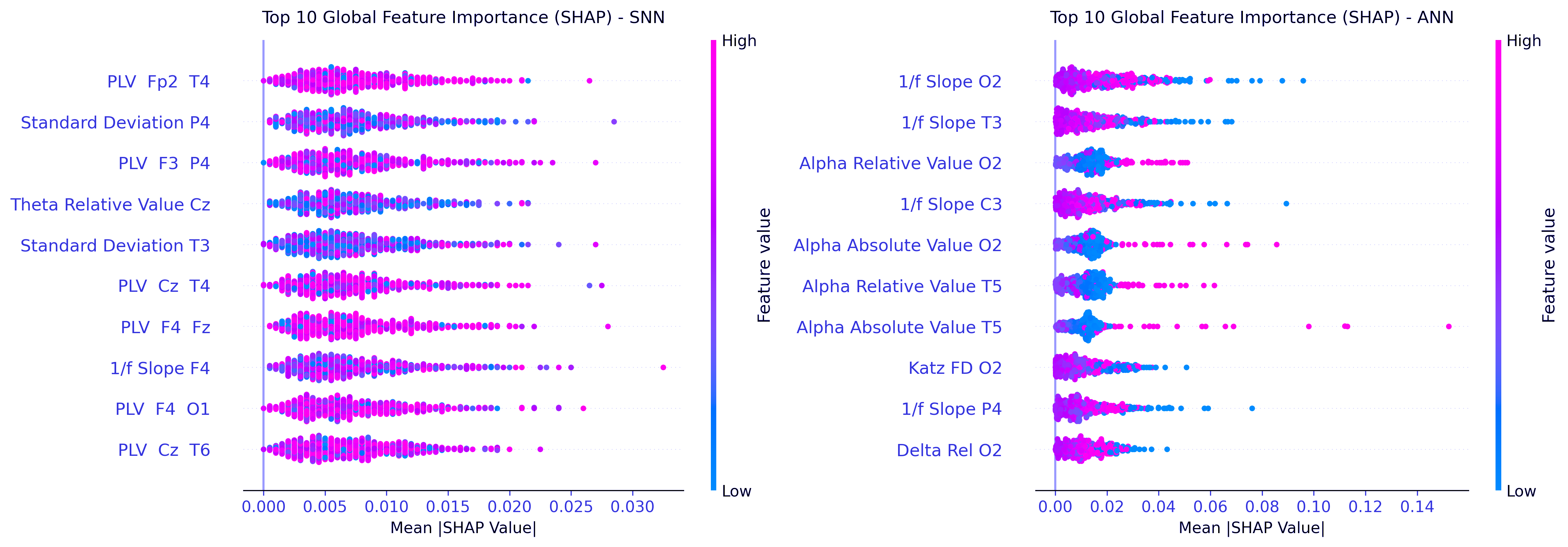}
    \caption{Top 10  global feature relevance analysis of the  SNN and ANN using SHAP. Each subfigure represents the mean SHAP values across EEG windows for SNN (left) and ANN (right). Each dot represents 
one EEG sample / epoch / subject evaluated for that feature. 200 Multiple EEG-derived features were used, including spectral bands, connectivity metrics, standard deviation (signal amplitude variability) and the inverse power-law slope of the EEG spectrum. }
    \label{fig:shap_snn}
\end{figure*}

Figure \ref{AUC} shows the AUROC for both the proposed SNN  and corresponding ANN classifier, while Figure \ref{fig:shap_snn} shows the relevant features used respectively for the SNN and ANN classifier. Here we show that the 1/f slope is one of the top 10  relevant features for classification among all 200. 

\subsection{Network based statistics}
\begin{figure*}[t]
\centering

\begin{minipage}[t]{0.78\textwidth}
\centering

\begin{subfigure}[t]{0.19\textwidth}
  \centering
  \includegraphics[width=\linewidth]{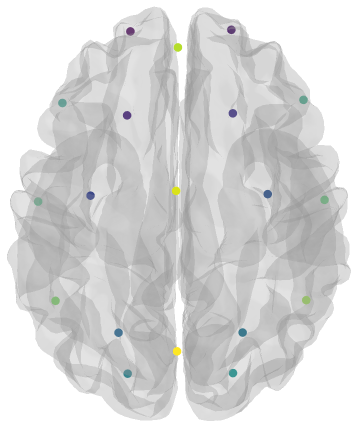}
  \caption{}
\end{subfigure}
\begin{subfigure}[t]{0.19\textwidth}
  \centering
  \includegraphics[width=\linewidth]{nocon.png}
  \caption{}
\end{subfigure}
\begin{subfigure}[t]{0.19\textwidth}
  \centering
  \includegraphics[width=\linewidth]{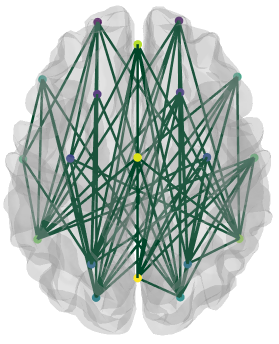}
  \caption{}
\end{subfigure}
\begin{subfigure}[t]{0.19\textwidth}
  \centering
  \includegraphics[width=\linewidth]{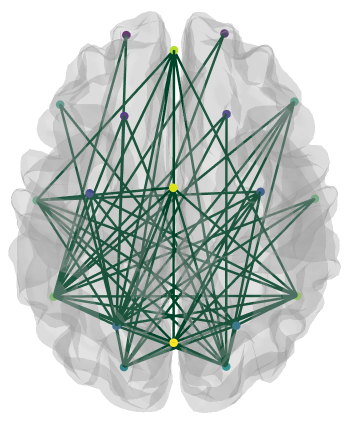}
  \caption{}
\end{subfigure}
\begin{subfigure}[t]{0.19\textwidth}
  \centering
  \includegraphics[width=\linewidth]{nocon.png}
  \caption{}
\end{subfigure}

\vspace{0.5em}

\begin{subfigure}[t]{0.19\textwidth}
  \centering
  \includegraphics[width=\linewidth]{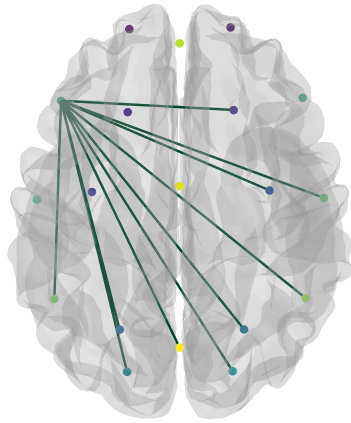}
  \caption{}
\end{subfigure}
\begin{subfigure}[t]{0.19\textwidth}
  \centering
  \includegraphics[width=\linewidth]{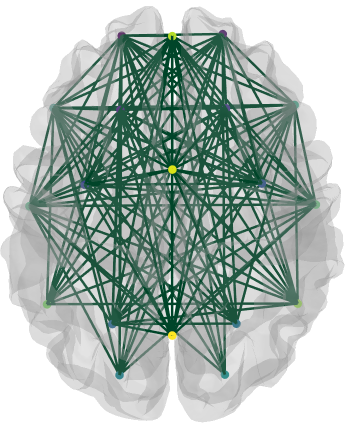}
  \caption{}
\end{subfigure}
\begin{subfigure}[t]{0.19\textwidth}
  \centering
  \includegraphics[width=\linewidth]{nocon.png}
  \caption{}
\end{subfigure}
\begin{subfigure}[t]{0.19\textwidth}
  \centering
  \includegraphics[width=\linewidth]{nocon.png}
  \caption{}
\end{subfigure}
\begin{subfigure}[t]{0.19\textwidth}
  \centering
  \includegraphics[width=\linewidth]{nocon.png}
  \caption{}
\end{subfigure}

\end{minipage}
\hfill
\begin{minipage}[c]{0.2\textwidth}
\centering
\begin{subfigure}[t]{\linewidth}
  \centering
\includegraphics[height=0.2\textheight]{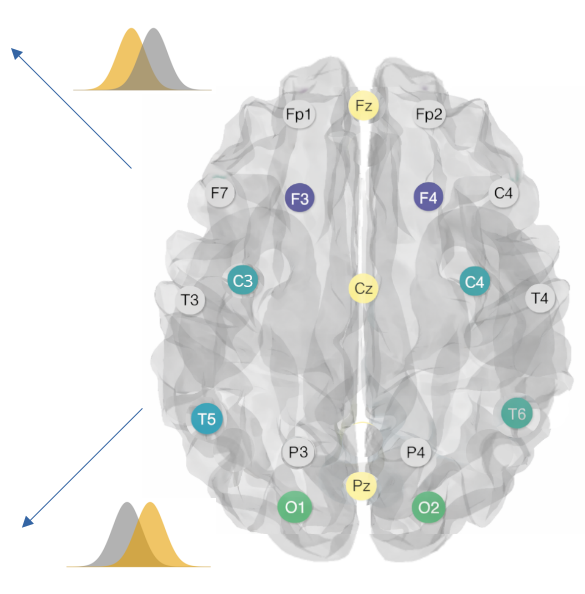}
  \caption{}
\end{subfigure}
\end{minipage}

\caption{Significant functional connectivity edges identified by the NBS analysis depicted in axial views. Subfigures (a–e) show the significant connections for the left-tailed NBS contrast in the delta, theta, alpha, beta, and gamma frequency bands, respectively. Subfigures (f–j) report the corresponding results for the right-tailed NBS contrast across the same frequency bands. Each subfigure displays only the edges belonging to the significant NBS component detected for that frequency band; subfigures with no visible edges indicate that no significant component was found. Subfigure (k) shows the EEG electrode layout, provided as a spatial reference for interpreting the connectivity patterns.}
\label{fig:NBS}
\end{figure*}

The NBS analysis led to some statistically 
significant results for some bands while nothing was detected for others. The total number of discriminant edges for each band is reported in Table \ref{tab:nbs_edges}, while the individual discriminant edges are reported in Figure  \ref{fig:NBS}. The statistically significant edges are also reported in two rows to differentiate the two tails of the test: 
In the first row, the left-tailed tests correspond to decreased connectivity in AD (HC $>$ AD), while in the second row the right-tailed tests correspond to increased connectivity in AD (AD $>$ HC). The reference for the channels depicted in Figure \ref{fig:NBS}k.

\begin{table}[!t]
\centering
\caption{Number of significant connections identified by NBS across frequency bands. Left-tailed tests correspond to decreased connectivity in AD (HC $>$ AD), while right-tailed tests correspond to increased connectivity in AD (AD $>$ HC).}
\label{tab:nbs_edges}
\begin{tabular}{lcc}
\hline
\textbf{Frequency Band} & \textbf{Left-tailed (HC $>$ AD)} & \textbf{Right-tailed (AD $>$ HC)} \\
\hline
$\delta$ (0.5--4) Hz   & 0   & 20 \\
$\theta$ (4--8) Hz  & 0   & 260 \\
$\alpha$ (8--13)  Hz& 146 & 0   \\
$\beta$ (13--30)  Hz& 128 & 0   \\
$\gamma$ (30--45)  Hz& 0   & 0   \\
\hline
\end{tabular}
\end{table}

\subsection{Simulations}
\begin{figure}[!]
  \centering
  \includegraphics[width=0.99 \linewidth]{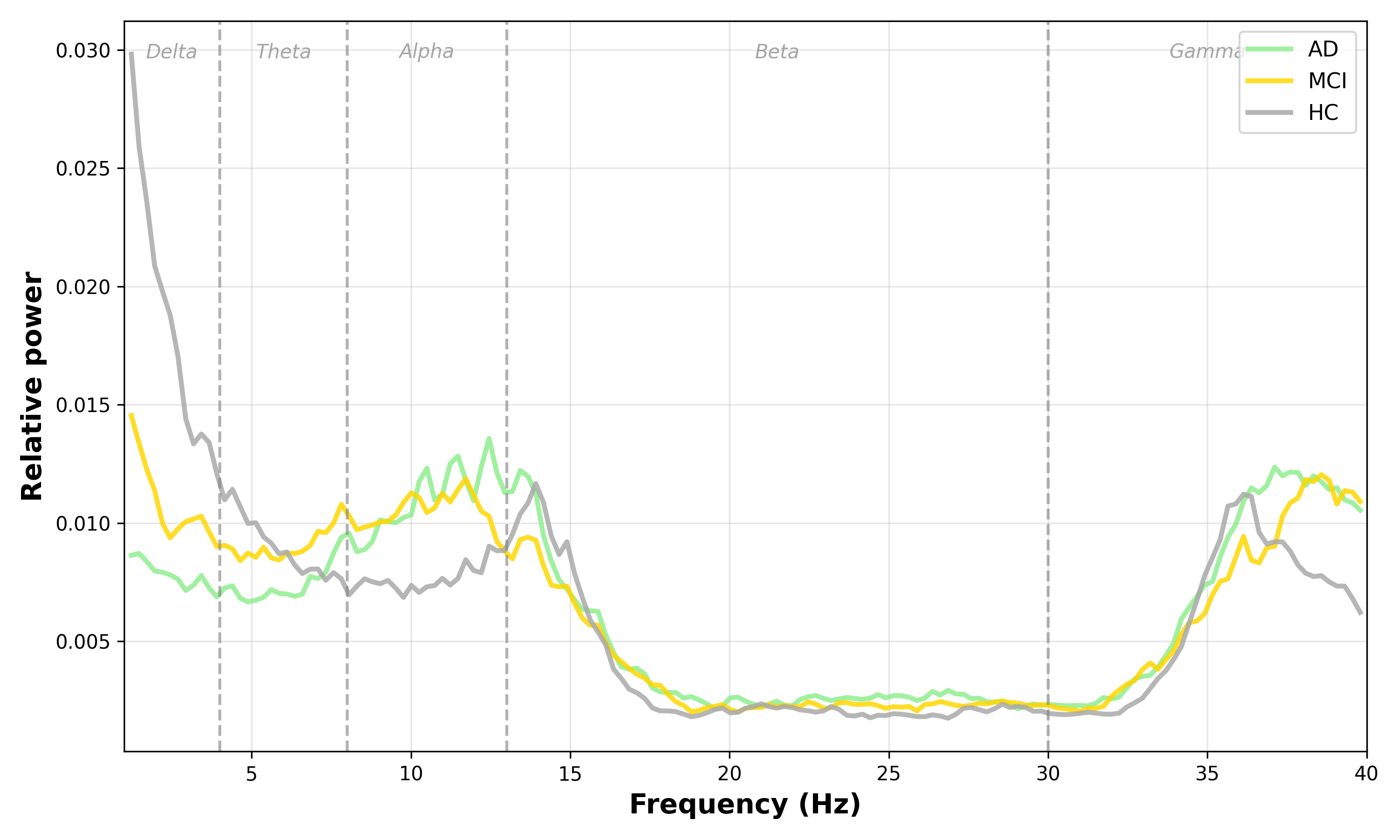}
  \caption{ Averaged relative power for Model 1 without FC-informed structure.}
  \label{fig:raw_PSD}
\end{figure}

\begin{figure}[!]
  \centering
  \includegraphics[width=0.99\linewidth]{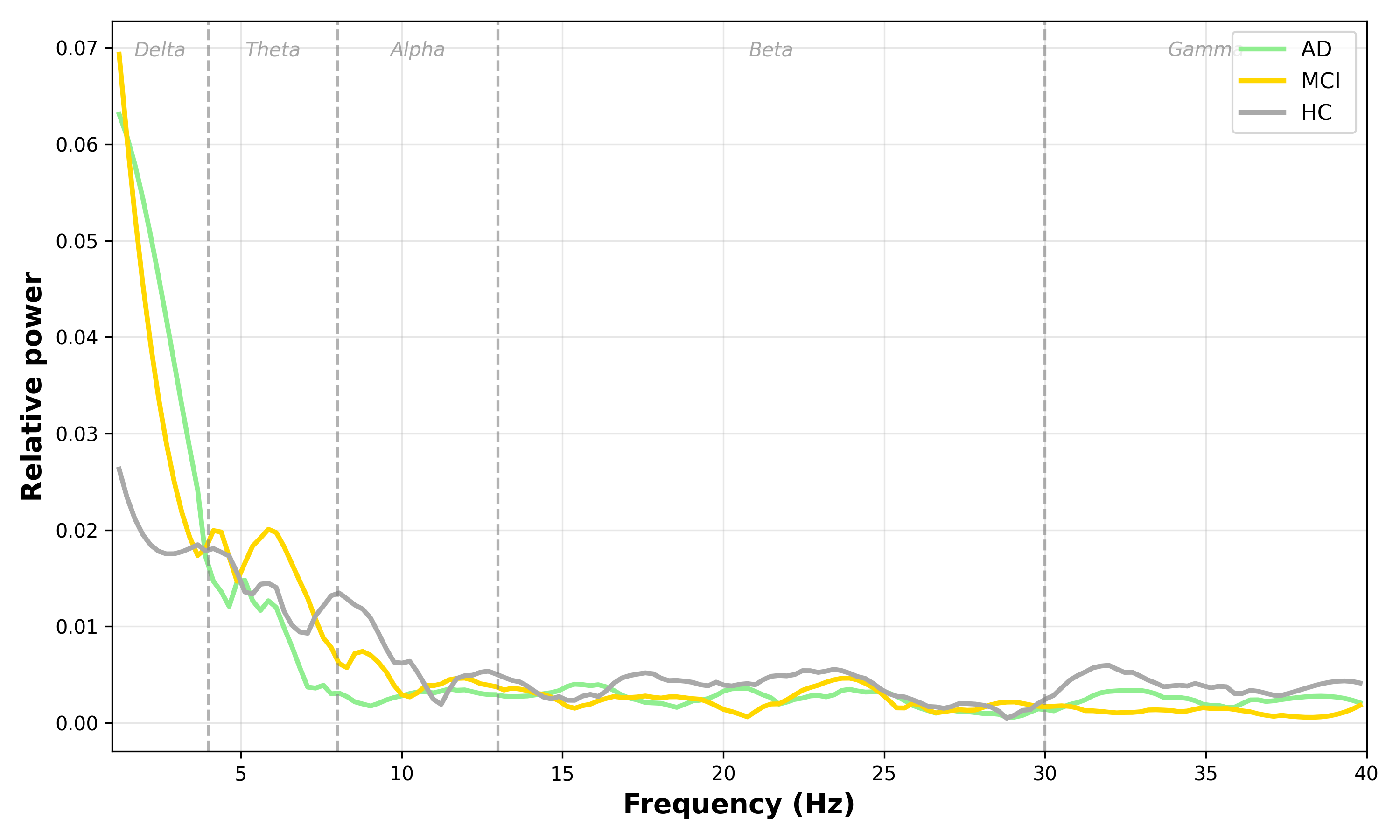}
  \caption{ Averaged relative power for Model 2 without FC-informed structure.}
  \label{fig:PSD}
\end{figure}

\begin{figure}[!]
  \centering
  \includegraphics[width=0.99 \linewidth]{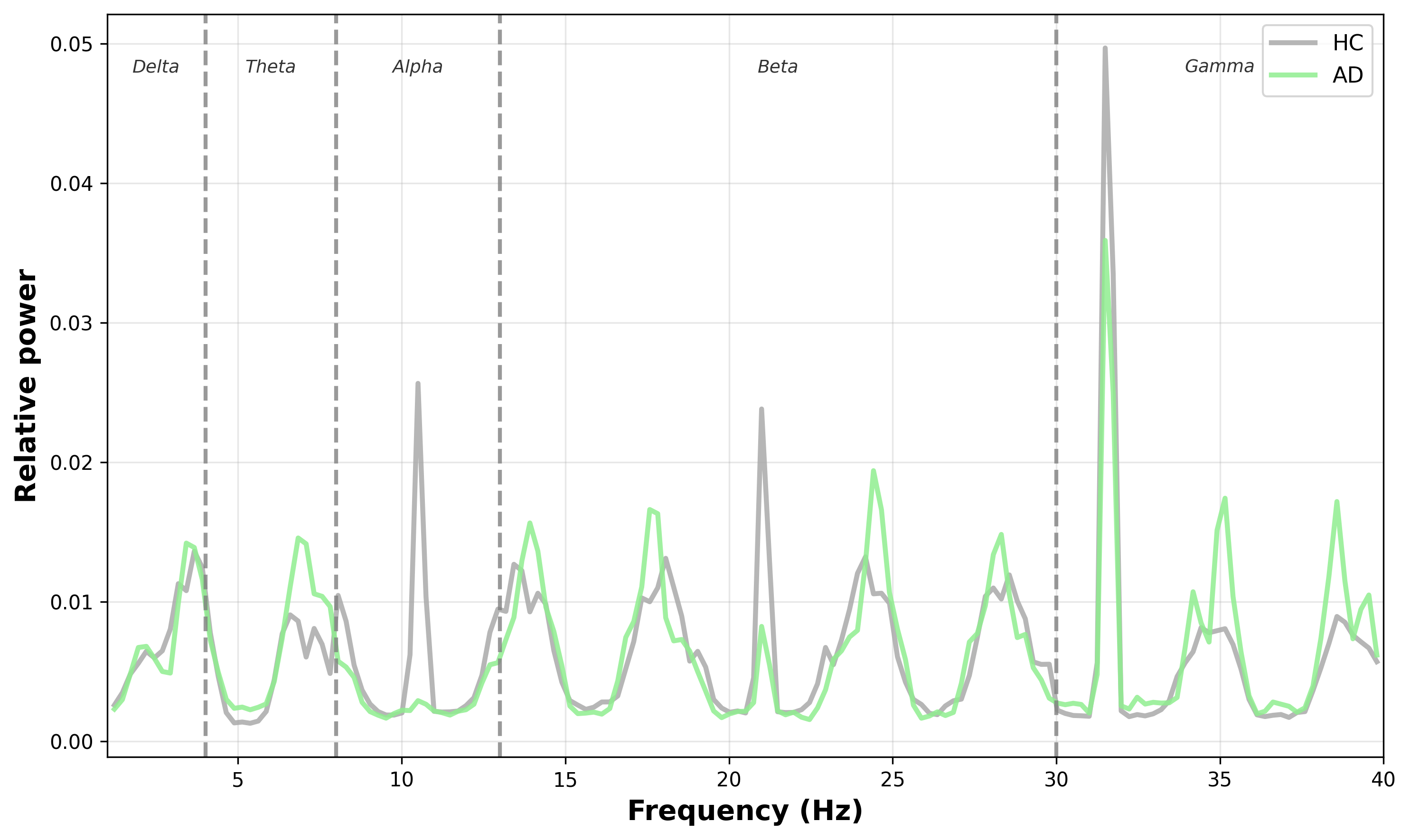}
  \caption{Averaged relative power for Model 1 with multi-network simulation constrained by human FC. }
  \label{fig:FC_PSD}
\end{figure}

\begin{figure}[!]
  \centering
  \includegraphics[width=0.99 \linewidth]{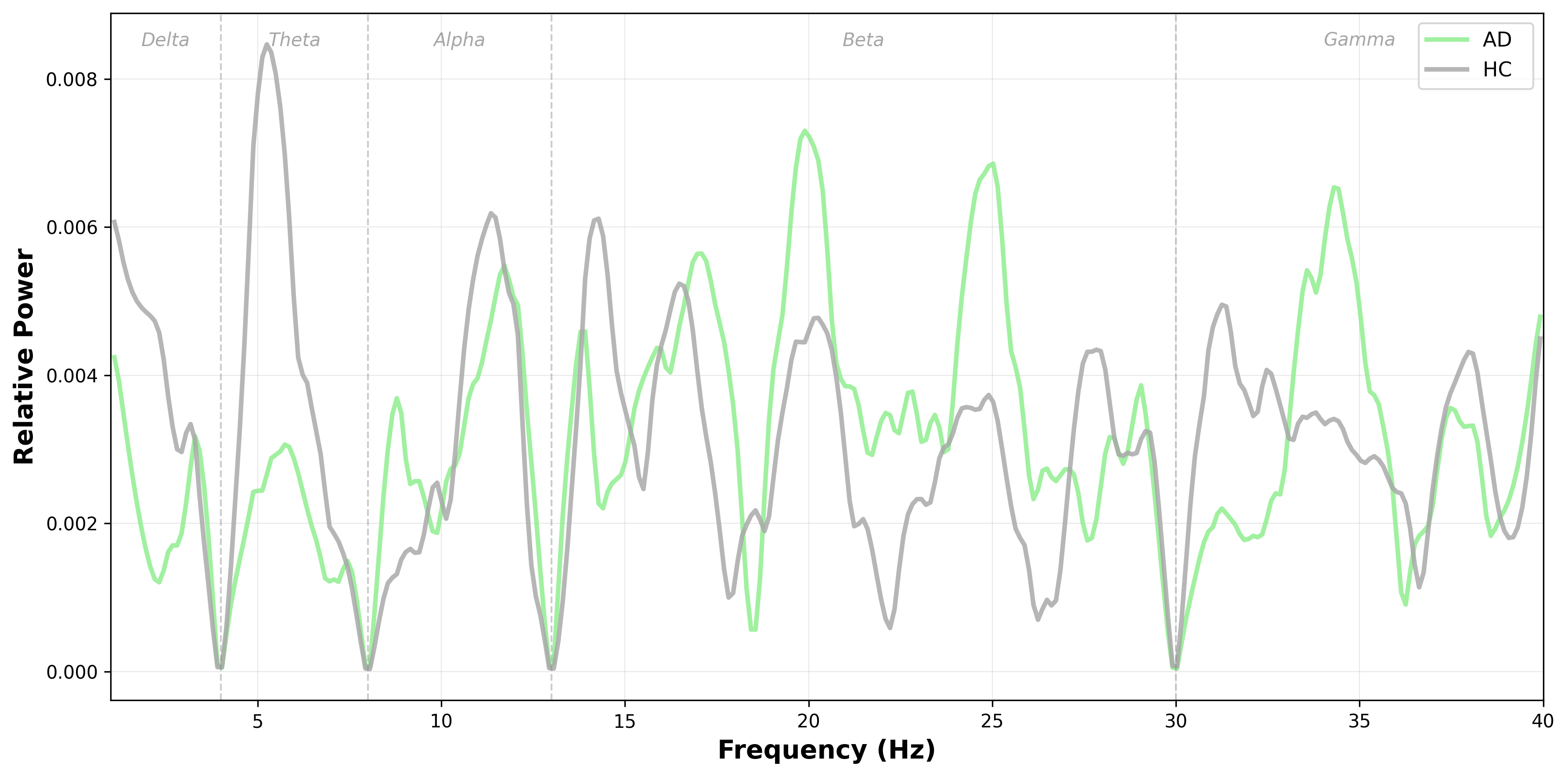}
  \caption{Averaged relative power for Model 2 with multi-network simulation constrained by human FC. }
  \label{fig:FC_PSD2}
\end{figure}

The power spectrum for the two simulation is reported in Figure \ref{fig:raw_PSD} and \ref{fig:PSD}, respectively for the potential-based proxy and the synaptic current-based EEG proxy. The FC-informed simulations again for the two proxys are instead depicted in Figure \ref{fig:FC_PSD} and \ref{fig:FC_PSD2}.  Table \ref{cohens} reports 
the comparison of Cohen's $d$, and other summary statistics for different models and empirical EEG data, analyzing the aperiodic exponent (1/f slope). The column $|\Delta d|$ shows the absolute difference between the model Cohen's $d$ and the EEG Cohen's $d$. 
All simulations were performed on a standard laptop (Intel Core Ultra 5 CPU, 16 GB RAM). Individual runs of Model 1 and Model 2 required approximately 1 minute without functional connectivity priors and approximatively 5 minutes when FC priors were included. 

\begin{table}[h!]
\centering
\caption{Comparison of Cohen's $d$, pooled standard deviation, and group means for different models and empirical EEG data, analyzing the aperiodic exponent (1/f slope). The column $|\Delta d|$ shows the absolute difference between the model Cohen's $d$ and the EEG Cohen's $d$.} \label{cohens}
\begin{tabular}{lccccc}
\hline
\textbf{Case} & \textbf{Cohen's $d$} & \textbf{$|\Delta d|$} & \textbf{$s_\text{pooled}$} & \textbf{Mean AD} & \textbf{Mean HC} \\
\hline
M. 1 & -1.332 & 1.185 & 0.242 & 0.430 & 0.752 \\
M. 2 & -1.324 & 1.177 & 0.301 & -0.582 & -0.183 \\
M. 1 FC-based& -0.853 & 0.706 & 0.125 & -0.199 & -0.093 \\
M. 2 FC-based& -0.190 & 0.043 & 0.430 & -0.110 & -0.030 \\
EEG analysis & -0.147 & -  & 0.146 & 2.166 & 2.187 \\
\hline
\end{tabular}
\end{table}

\section{Discussions}
The search for computational biomarkers for AD, which are  mechanistically interpretable, represents a major challenge in computational neurology. This study addresses this challenge   by relating two typically disparate approaches: a learning-based, energy-efficient spiking neural network (SNN) classifier for EEG analysis, and biophysical simulations of cortical microcircuits, forming a neuro-bridge between two generally different points of view. We demonstrate that the aperiodic slope of the EEG power spectrum, a putative marker of excitation-inhibition (E/I) balance, can be used  as a  feature for SNN-based classification of AD. By modeling this E/I imbalance directly in spiking networks, we show that the spectral consequences of altered inhibitory control spectral slowing and reduced alpha coherence, and closely resemble the empirical EEG phenotypes of AD.

From a classification standpoint, both the SNN and ANN models achieved robust discrimination between AD and HC EEG, with the ANN reaching a higher AUROC (0.936) compared to the SNN (0.839). While the ANN demonstrates superior predictive performance, the SNN attains competitive accuracy despite its constrained, event-driven architecture and lower representational capacity. Importantly, the reduced performance of the SNN does not indicate a lack of sensitivity to disease-relevant information, but rather reflects a trade-off between classification optimality and biophysical plausibility.

SHAP analysis revealed that the most informative features for both models are neurophysiologically meaningful and largely overlap, including  PLV  measures in fronto-temporal and centro-parietal networks, band-limited relative power (particularly alpha, beta, and theta), signal variability (standard deviation), and the aperiodic spectral slope. These features are closely related to neuronal synchronization and excitation–inhibition balance, reinforcing their relevance as mechanistic biomarkers of AD. Notably, connectivity-based features dominate feature importance rankings, consistent with the hypothesis that AD is characterized by disrupted large-scale coordination rather than isolated spectral changes, which is further confirmed by the NBS analysis.  
The SHAP analysis indicates that the ANN predominantly exploits aperiodic spectral features, particularly the 1/f slope, which has been directly linked to excitation–inhibition balance in prior work. This aligns closely with the primary neurobiological hypothesis of the present study. In contrast, the SNN places greater emphasis on connectivity and variability measures, which do not constitute direct biomarkers of E/I balance but instead reflect network-level coordination and temporal dynamics. 
Accordingly, the two models play distinct and complementary roles: the ANN is well suited for detecting E/I-related spectral signatures in empirical EEG data, whereas the SNN provides a mechanistic modeling framework in which such signatures can be generated, perturbed, and interpreted in terms of underlying spiking circuit dynamics. The slightly lower AUROC of the SNN thus reflects a deliberate emphasis on biological realizability and dynamical interpretability rather than maximal biomarker sensitivity.

EEG has long been used to characterize functional brain alterations in AD, revealing consistent frequency-specific changes that distinguish patients from HC. These alterations reflect disrupted neuronal synchronization, synaptic loss, and large-scale network disorganization associated with neurodegeneration \cite{jeong2004eeg}.  
An increase in theta and delta activities and a decrease in alpha and beta activities are repeatedly observed \cite{brenner1986computerized ,
coben1985longitudinal, giaquinto1986eeg}. Those patterns are consistent with the NBS results. 
Findings in the gamma band are less consistent than those observed at lower frequencies,
They are present in Alzheimer’s disease but show greater variability across studies  \cite{simfukwe2025investigating}. Indeed, in our NBS experiments we did not find any statistically significant connections for this band, though some parts of the spectrum for the simulations it shows some differences.
This incongruence suggests that gamma-related alterations may be present but remain subthreshold or spatially fragmented, thus not forming large connected components detectable by NBS. This interpretation aligns with prior work indicating that gamma abnormalities in Alzheimer’s disease are more reliably captured using task-based paradigms, cross-frequency coupling, or local circuit measures rather than large-scale resting-state connectivity.

In agreement with literature, we found differences for the delta band, though the discriminant connections seem to be all related to the F7 channel. This limited result might reflect the aforementioned limitations of using the delta band for PLV estimators, since those might be biased and have a larger variance in the low frequency bands and can give a noisy estimator \cite{vinck2010pairwise}.  

Regarding the simulations, 
 alterations in the E/I balance produce consistent across the models and interpretable changes in low-frequency bands, particularly delta, theta, and alpha. Importantly, both models reproduce a shift towards lower-frequency dominance under altered E/I regimes, aligning with the well-established slowing of EEG observed in Alzheimer’s disease and related conditions.

In Model 1, changes in relative power are broadly distributed across frequencies, with noticeable modulation in the delta/theta/alpha range though appearing in almost total disagreement with the NBS results. 
Model~2 (EEG-proxy), by contrast, exhibits strong attenuation of beta power across conditions, resulting in an even flatter spectrum between $\sim15–30$ Hz compared to Model 1. This difference highlights how beta-band activity may be less robust at the level of far-field EEG proxies, particularly in simplified homogeneous networks lacking structured long-range coupling. Moreover, in Model 1, the membrane-potential–based network signal exhibits visible fluctuations and condition-dependent differences in the gamma range, suggesting that local circuit dynamics and fast excitatory–inhibitory interactions can generate high-frequency components. These gamma-range effects, however, appear fragmented and spatially local, consistent with high-frequency activity being driven by microcircuit mechanisms involving fast-spiking interneurons.

This might indicate that while the contributions of the gamma-band to the population-averaged synaptic current remain smaller than the low-frequency components, they are not fully suppressed at the macroscopic EEG level. 
However, the limited magnitude of these differences implies that gamma activity contributes weakly to large-scale network formation, helping to explain why local circuit-level gamma alterations may not reliably translate into robust EEG-based connectivity effects detectable by methods such as NBS.

Regarding the empirically derived functional connectivity (FC) informing  the simulations,
we can notice similar behaviors to the other simulations, although  compared to Models 1 and 2, the FC-informed simulations show clearer and more stable spectral separation between HC and AD conditions in the beta and alpha ranges with visible peaks. In particular, alpha-band differences appear sharper and more spatially coherent, consistent with the idea that alpha oscillations are strongly shaped by large-scale cortical networks rather than purely local circuit dynamics. The difference in the delta-band between the two models  occurs because Model 1 measures slow single-neuron membrane potential fluctuations, which grow with stronger inhibition, whereas Model 2 measures population-averaged synaptic currents, where stronger inhibition suppresses slow collective fluctuations leading to opposite low-frequency trends. A comparison previously reported in the literature \cite{deco2018whole}. Moreover, the spectrum of these analysis is given by stitching simulation runs using different functional connectivity, which might introduce those aspects. 

By incorporating empirically derived band-specific functional connectivity priors into multi-subnetwork simulations, the differentiation between AD and healthy conditions in the theta and alpha bands was further enhanced. This result underscores the importance of large-scale network topology in shaping macroscopic EEG rhythms and suggests that functional connectivity can constitute a stronger prior than E/I balance alone to explain disease-related spectral organization at the systems level. At the same time, the persistence of weak gamma effects in FC-informed simulations suggests the view that resting-state gamma alterations in AD are spatially fragmented and poorly captured by large-scale connectivity analyzes. Previous 
works found a significant difference in the alpha band with higher relative power for the HC series compared to the AD series \cite{MartinezCanada2023, meghdadi2021resting,diachenko2024functional}. This behavior is  visible in our simulations in Model 1 FC-inforced and  Model 2 without FC prior. Indeed, comparing the graphs qualitatively, Model 2 without prior FC resembles more previous results with EEG data \cite{MartinezCanada2023, meghdadi2021resting, diachenko2024functional}. 

Looking at quantitative relationships between the simulations and the 1/f slope obtained by the empirical EEG of human data, we observed a consistency in the difference between the two populations AD vs HC, with also a consistent sign comparing Cohen's $d$ scores. 
With respect to magnitude, the behaviors were different.  Our simulations revealed that models without functional connectivity (Model 1 and Model 2) produced large effect sizes for the aperiodic exponent (Cohen’s $d \approx -1.33$), substantially exceeding those observed in the empirical EEG data (Cohen’s $d \approx -0.15$). Introducing functional connectivity constraints (Model 1 FC-based and Model 2 FC-based) reduced the effect sizes, with Model 2 FC-based approaching the empirical EEG magnitude ($|\Delta d| \approx 0.04$), while preserving the correct direction of change. These results indicate that, while excitatory/inhibitory imbalance strongly affects neuronal dynamics in the model, the translation to scalp-level EEG is attenuated. Several factors likely contribute to this discrepancy: EEG primarily reflects summed cortical activity, is insensitive to subcortical dynamics, and is subject to noise and volume conduction, which reduce observable effect sizes. The comparison highlights the importance of functional network interactions in constraining network dynamics and bridging the gap between neuronal-level changes and EEG signatures, and suggests that random networks alone may overestimate the impact of excitatory imbalance when projected to the EEG scale. Therefore, we might conclude that Model 2 without prior is the behaviour closer to what was expected, but the FC-inforced models are closer to reality due to their high variability and attenuated effect. 

\subsection{Limitations of the study}
Several limitations of the present study should be acknowledged. First, our simulations relied on randomly connected networks rather than detailed laminar cortical models, which limits the ability to capture layer-specific dynamics and microcircuit organization. Even so, as reported in  the literature, this is generally accepted as it can already provide a more direct measure, and laminar structures increase complexity with often little benefits in this context \cite{Brunel2000,vogels2005signal,MartinezCanada2023, gao2017inferring}.  
Second, empirical EEG provides only a coarse measure of cortical activity and represents a relatively weak prior compared to structural or anatomical constraints, reducing sensitivity to subtle network changes, and even more caution should be given to the fact we have a small sample size available. Indeed, we wanted to investigate relative power changes and not to propose a realistic model of brain simulation, otherwise also a model using millions of synthetic neurons would be necessary.  
Finally, in this same context, subcortical contributions, such as thalamic input, are generally not captured in the EEG recordings, preventing the study of interactions between  the cortical and subcortical regions that are known to shape oscillatory dynamics. Together, these factors likely contribute to the observed discrepancies between model predictions and scalp-level EEG signatures, and suggest  future work that incorporates laminar structure \cite{Kusch2024MultiscaleCoSimulation}, structural priors, and subcortical signals may provide a more complete understanding of excitatory/inhibitory imbalance in Alzheimer’s disease. 
Nevertheless, our results suggest that neuromorphic computing, grounded in biophysical realism, offers a promising pathway toward interpretable tools for large-scale neurological screening and monitoring.

\section{Conclusions}
In this work, we presented a unified neuromorphic and mechanistic framework for understanding and detecting neurodegeneration from resting-state EEG. By combining spiking neural network   classification with biophysically grounded spiking network simulations, we linked data-driven EEG biomarkers to underlying cortical  E/I imbalance, offering both predictive performance and mechanistic interpretability.

From a learning perspective, our results demonstrate that SNNs can successfully discriminate AD patients from healthy controls using EEG-derived features, with particular sensitivity to aperiodic spectral characteristics such as the inverse power-law slope. This finding supports the hypothesis that aperiodic EEG components encode information about large-scale E/I balance and neurodegenerative processes, and shows that neuromorphic models can exploit these signatures efficiently. Importantly, the SNN architecture achieves this while remaining compatible with low-power, event-driven computation, highlighting its potential for scalable and deployable EEG-based diagnostic systems.

From a mechanistic standpoint, our spiking network simulations reveal that systematic manipulation of the inhibitory-to-excitatory synaptic strength ratio reproduces key EEG phenotypes observed in AD, including spectral slowing, altered alpha organization, and reduced higher-frequency coherence. By comparing the membrane-potential–based network signal and the synaptic-current–based network signal,  we showed that low- and mid-frequency alterations robustly propagate from local circuit dynamics to macroscopic signals, whereas gamma-band effects tend to remain confined to microcircuit activity and are strongly attenuated at the scalp EEG level. This multiscale dissociation provides a principled explanation for the absence of statistically significant gamma-band components in our NBS analysis, despite subtle gamma-related effects observed in local simulations.

Taken together, these findings support a coherent multilevel interpretation of EEG alterations in Alzheimer’s disease: E/I imbalance acts as a microcircuit-level driver of spectral changes, while large-scale functional connectivity determines which of these alterations survive spatial averaging and emerge as robust macroscopic biomarkers. By explicitly linking neuromorphic learning, network-based statistics, and mechanistic simulations, this study goes beyond black-box classification and provides a guiding framework for interpreting neurodegeneration EEG biomarkers.

Although E/I imbalance models do not yield a direct cure for Alzheimer’s disease, they identify mechanisms through which network dysfunction arises and highlight targets for stabilizing brain activity before irreversible damage occurs, including GABAergic interneuron dysfunction, altered synaptic gain control, and disrupted NMDA/AMPA–GABA coupling \cite{palop2010amyloid}. These insights support therapeutic strategies aimed at restoring network balance, such as low-dose antiepileptics, selective GABAergic receptor modulation, and non-invasive neuromodulation targeting impaired oscillatory rhythms \cite{ebrahimi2025glutamatergic}.

\section*{Acknowledgment}
The authors thank Elisa Donati and Chiara DeLuca 
for useful insights and conversations.

\bibliographystyle{IEEEtran}

\bibliography{neuro}

@article{Brunel2000,
  author    = {Brunel, Nicolas},
  title     = {Dynamics of sparsely connected networks of excitatory and inhibitory spiking neurons},
  journal   = {Journal of Computational Neuroscience},
  year      = {2000},
  volume    = {8},
  number    = {3},
  pages     = {183--208},
  doi       = {10.1023/A:1008925309027},
  url       = {https://link.springer.com/article/10.1023/A:1008925309027}
}

@article{palop2010amyloid,
  title={Amyloid-$\beta$--induced neuronal dysfunction in {Alzheimer}'s disease: from synapses toward neural networks},
  author={Palop, Jorge J and Mucke, Lennart},
  journal={Nature neuroscience},
  volume={13},
  number={7},
  pages={812--818},
  year={2010},
  publisher={Nature Publishing Group US New York}
}

@article{simfukwe2025investigating,
  title={Investigating Gamma Frequency Band PSD in Alzheimer’s Disease Using qEEG from Eyes-Open and Eyes-Closed Resting States},
  author={Simfukwe, Chanda and An, Seong Soo A and Youn, Young Chul},
  journal={Journal of Clinical Medicine},
  volume={14},
  number={12},
  pages={4256},
  year={2025},
  publisher={Multidisciplinary Digital Publishing Institute}
}

@article{dauwels2010diagnosis,
  title={Diagnosis of Alzheimer's disease from EEG signals: where are we standing?},
  author={Dauwels, Justin and Vialatte, Francois and Cichocki, Andrzej},
  journal={Current Alzheimer Research},
  volume={7},
  number={6},
  pages={487--505},
  year={2010},
  publisher={Bentham Science Publishers direct}
}

@article{ebrahimi2025glutamatergic,
  title={Glutamatergic and GABAergic metabolite levels in {Alzheimer}’s disease: a systematic review and meta-analysis},
  author={Ebrahimi, Rasoul and Mohammad Soltani, Sana and Masouri, Mohammad Mahdi and Seifi, Mojtaba and Ghafourian, Kiana and Noori, Shokoofe},
  journal={BMC neurology},
  volume={25},
  number={1},
  pages={344},
  year={2025},
  publisher={Springer}
}

@article{deco2018whole,
  title={Whole-brain multimodal neuroimaging model using serotonin receptor maps explains non-linear functional effects of LSD},
  author={Deco, Gustavo and Cruzat, Josephine and Cabral, Joana and Knudsen, Gitte M and Carhart-Harris, Robin L and Whybrow, Peter C and Logothetis, Nikos K and Kringelbach, Morten L},
  journal={Current biology},
  volume={28},
  number={19},
  pages={3065--3074},
  year={2018},
  publisher={Elsevier}
}

@article{palop2016network,
  title={Network abnormalities and interneuron dysfunction in {Alzheimer} disease},
  author={Palop, Jorge J and Mucke, Lennart},
  journal={Nature Reviews Neuroscience},
  volume={17},
  number={12},
  pages={777--792},
  year={2016},
  publisher={Nature Publishing Group UK London}
}

@article{diachenko2024functional,
  title={Functional excitation-inhibition ratio indicates near-critical oscillations across frequencies},
  author={Diachenko, Marina and Sharma, Additya and Smit, Dirk JA and Mansvelder, Huibert D and Bruining, Hilgo and De Geus, Eco and Avramiea, Arthur-Ervin and Linkenkaer-Hansen, Klaus},
  journal={Imaging Neuroscience},
  volume={2},
  pages={1--17},
  year={2024},
  publisher={MIT Press 255 Main Street, 9th Floor, Cambridge, Massachusetts 02142, USA~…}
}

@article{van2023resting,
  title={Resting-state oscillations reveal disturbed excitation--inhibition ratio in {Alzheimer}’s disease patients},
  author={Van Nifterick, Anne M and Mulder, Danique and Duineveld, Denise J and Diachenko, Marina and Scheltens, Philip and Stam, Cornelis J and Van Kesteren, Ronald E and Linkenkaer-Hansen, Klaus and Hillebrand, Arjan and Gouw, Alida A},
  journal={Scientific reports},
  volume={13},
  number={1},
  pages={7419},
  year={2023},
  publisher={Nature Publishing Group UK London}
}

@article{jeong2004eeg,
  title={{EEG} dynamics in patients with {Alzheimer}'s disease},
  author={Jeong, Jaeseung},
  journal={Clinical Neurophysiology},
  volume={115},
  number={7},
  pages={1490--1505},
  year={2004},
  publisher={Elsevier}
}

@article{brenner1986computerized,
  title={Computerized {EEG} spectral analysis in elderly normal, demented and depressed subjects},
  author={Brenner, R.P. and Ulrich, R.F. and Spiker, D.G. and Sclabassi, R.J. and Reynolds, C.F. and Marin, R.S. and Boller, F.},
  journal={Electroencephalography and Clinical Neurophysiology},
  volume={64},
  number={6},
  pages={483--492},
  year={1986},
  publisher={Elsevier}
}

@article{coben1985longitudinal,
  title={A longitudinal {EEG} study of mild senile dementia of {Alzheimer} type: changes at 1 year and at 2.5 years},
  author={Coben, L.A. and Danziger, W. and Storandt, M.},
  journal={Electroencephalography and Clinical Neurophysiology},
  volume={61},
  number={2},
  pages={101--112},
  year={1985},
  publisher={Elsevier}
}

@article{giaquinto1986eeg,
  title={The {EEG} in the normal elderly: a contribution to the interpretation of aging and dementia},
  author={Giaquinto, S. and Nolfe, G.},
  journal={Electroencephalography and Clinical Neurophysiology},
  volume={63},
  number={6},
  pages={540--546},
  year={1986},
  publisher={Elsevier}
}

@article{Zalesky2010NBS,
  author  = {Zalesky, Andrew and Fornito, Alex and Bullmore, Edward T.},
  title   = {Network‐based statistic: identifying differences in brain networks},
  journal = {NeuroImage},
  volume  = {53},
  number  = {4},
  pages   = {1197--1207},
  year    = {2010},
  doi     = {10.1016/j.neuroimage.2010.06.041}
}

@article{al2024disrupted,
  title={Disrupted brain functional connectivity as early signature in cognitively healthy individuals with pathological CSF amyloid/tau},
  author={Al-Ezzi, Abdulhakim and Arechavala, Rebecca J and Butler, Ryan and Nolty, Anne and Kang, Jimmy J and Shimojo, Shinsuke and Wu, Daw-An and Fonteh, Alfred N and Kleinman, Michael T and Kloner, Robert A and others},
  journal={Communications Biology},
  volume={7},
  number={1},
  pages={1037},
  year={2024},
  publisher={Nature Publishing Group UK London}
}

@article{vinck2010pairwise,
  title={The pairwise phase consistency: A bias-free measure of rhythmic neuronal synchronization},
  author={Vinck, Martin and van Wingerden, Marijn and Womelsdorf, Thilo and Fries, Pascal and Pennartz, Cyriel M.A.},
  journal={NeuroImage},
  volume={51},
  number={1},
  pages={112--122},
  year={2010},
  doi={10.1016/j.neuroimage.2010.01.073}
}

@article{rojas2018study,
  title={{Study of resting-state functional connectivity networks using {EEG} electrodes position as seed}},
  author={Rojas, Gonzalo M and Alvarez, Carolina and Montoya, Carlos E and De la Iglesia-Vaya, MariA and Cisternas, Jaime E and G{\'a}lvez, Marcelo},
  journal={Frontiers in neuroscience},
  volume={12},
  pages={235},
  year={2018},
  publisher={Frontiers Media SA}
}

@article{miltiadous2023dataset,
  title={A dataset of scalp {EEG} recordings of {Alzheimer}’s disease, frontotemporal dementia and healthy subjects from routine {EEG}},
  author={Miltiadous, Andreas and Tzimourta, Katerina D and Afrantou, Theodora and Ioannidis, Panagiotis and Grigoriadis, Nikolaos and Tsalikakis, Dimitrios G and Angelidis, Pantelis and Tsipouras, Markos G and Glavas, Euripidis and Giannakeas, Nikolaos and others},
  journal={Data},
  volume={8},
  number={6},
  pages={95},
  year={2023},
  publisher={MDPI}
}

@article{blanco2024investigating,
  title={Investigating the interaction between {EEG} and fNIRS: A multimodal network analysis of brain connectivity},
  author={Blanco, Rosmary and Koba, Cemal and Crimi, Alessandro},
  journal={Journal of Computational Science},
  volume={82},
  pages={102416},
  year={2024},
  publisher={Elsevier}
}

@article{CAI2026108127,
title = {Spiking neural networks for {EEG} signal analysis: From theory to practice},
journal = {Neural Networks},
volume = {194},
pages = {108127},
year = {2026},
issn = {0893-6080},
doi = {https://doi.org/10.1016/j.neunet.2025.108127},
url = {https://www.sciencedirect.com/science/article/pii/S089360802501007X},
author = {Siqi Cai and Zheyuan Lin and Xiaoli Liu and Wenjie Wei and Shuai Wang and Malu Zhang and Tanja Schultz and Haizhou Li}
}

@article{Burelo2022,
author={Burelo, Karla
and Ramantani, Georgia
and Indiveri, Giacomo
and Sarnthein, Johannes},
title={A neuromorphic spiking neural network detects epileptic high frequency oscillations in the scalp {EEG}},
journal={Scientific Reports},
year={2022},
month={2},
day={02},
volume={12},
number={1},
pages={1798},
issn={2045-2322},
doi={10.1038/s41598-022-05883-8},
url={https://doi.org/10.1038/s41598-022-05883-8}
}

@article{Zhang2024,
author = {Zhang, Zongpeng and Xiao, Mingqing and Ji, Taoyun and Jiang, Yuwu and Lin, Tong and Zhou, Xiaohua and Lin, Zhouchen},
year = {2024},
month = {01},
pages = {1303564},
title = {Efficient and generalizable cross-patient epileptic seizure detection through a spiking neural network},
volume = {17},
journal = {Frontiers in Neuroscience},
doi = {10.3389/fnins.2023.1303564}
}

@article{LI2024109225,
title = {Real-time sub-milliwatt epilepsy detection implemented on a spiking neural network edge inference processor},
journal = {Computers in Biology and Medicine},
volume = {183},
pages = {109225},
year = {2024},
issn = {0010-4825},
doi = {https://doi.org/10.1016/j.compbiomed.2024.109225}, 
author = {Ruixin Li and Guoxu Zhao and Dylan Richard Muir and Yuya Ling and Karla Burelo and Mina Khoe and Dong Wang and Yannan Xing and Ning Qiao}
}

@article{Eppler2008PyNEST,
  author  = {Jochen Martin Eppler and Moritz Helias and Eilif Muller and Markus Diesmann and Marc-Oliver Gewaltig},
  title   = {PyNEST: a convenient interface to the NEST simulator},
  journal = {Frontiers in Neuroinformatics},
  year    = {2008},
  volume  = {2},
  pages   = {11},
  doi     = {10.3389/neuro.11.012.2008}
}

@article{MartinezCanada2023,
  author    = {Martínez-Cañada, Pablo and Perez-Valero, Elena and Minguillon, Juan and Pelayo, Francisco and López-Gordo, Miguel A. and Morillas, Carmen},
  title     = {Combining aperiodic 1/f slopes and brain simulation: An {EEG}/MEG proxy marker of excitation/inhibition imbalance in {Alzheimer}'s disease},
  journal   = {{Alzheimer}'s \& Dementia: Diagnosis, Assessment \& Disease Monitoring},
  year      = {2023},
  volume    = {15},
  number    = {3},
  pages     = {e12477},
  doi       = {10.1002/dad2.12477},
  url       = {https://doi.org/10.1002/dad2.12477}
}

@article{akbar2025unlocking,
  title={Unlocking the potential of {EEG} in {Alzheimer}'s disease research: Current status and pathways to precision detection},
  author={Akbar, Frnaz and Taj, Imran and Usman, Syed Muhammad and Imran, Ali Shariq and Khalid, Shehzad and Ihsan, Imran and Ali, Ammara and Yasin, Amanullah},
  journal={Brain Research Bulletin},
  pages={111281},
  year={2025},
  publisher={Elsevier}
}

@article{nichols2022estimation,
  title={Estimation of the global prevalence of dementia in 2019 and forecasted prevalence in 2050: an analysis for the Global Burden of Disease Study 2019},
  author={Nichols, Emma and Steinmetz, Jaimie D and Vollset, Stein Emil and Fukutaki, Kai and Chalek, Julian and Abd-Allah, Foad and Abdoli, Amir and Abualhasan, Ahmed and Abu-Gharbieh, Eman and Akram, Tayyaba Tayyaba and others},
  journal={The Lancet Public Health},
  volume={7},
  number={2},
  pages={e105--e125},
  year={2022},
  publisher={Elsevier}
}

@article{Kusch2024MultiscaleCoSimulation,
  title        = {Multiscale co‑simulation design pattern for neuroscience applications},
  author       = {Kusch, Lionel and Díaz‑Pier, Sandra and Klijn, Wouter and Sontheimer, Katrin and Bernard, Christine and Morrison, Abigail and Jirsa, Viktor K.},
  journal      = {Frontiers in Neuroinformatics},
  volume       = {18},
  year         = {2024},
  doi          = {10.3389/fninf.2024.1156683}
}

@article{donoghue2020parameterizing,
  title={Parameterizing neural power spectra into periodic and aperiodic components},
  author={Donoghue, Thomas and Haller, Matar and Peterson, Erik J and Varma, Paroma and Sebastian, Priyadarshini and Gao, Richard and Noto, Torben and Lara, Antonio H and Wallis, Joni D and Knight, Robert T and others},
  journal={Nature neuroscience},
  volume={23},
  number={12},
  pages={1655--1665},
  year={2020},
  publisher={Nature Publishing Group US New York}
}

@article{gao2017inferring,
  title={Inferring synaptic excitation/inhibition balance from field potentials},
  author={Gao, Richard and Peterson, Erik J and Voytek, Bradley},
  journal={Neuroimage},
  volume={158},
  pages={70--78},
  year={2017},
  publisher={Elsevier}
}

@article{lundberg2017unified,
  title={A unified approach to interpreting model predictions},
  author={Lundberg, Scott M and Lee, Su-In},
  journal={Advances in neural information processing systems},
  volume={30},
  year={2017}
}

@article{vogels2005signal,
  title={Signal propagation and logic gating in networks of integrate-and-fire neurons},
  author={Vogels, Tim P and Abbott, Larry F},
  journal={Journal of neuroscience},
  volume={25},
  number={46},
  pages={10786--10795},
  year={2005},
  publisher={Society for Neuroscience}
}

@article{mikhaylov2020neurohybrid,
  title={Neurohybrid memristive CMOS-integrated systems for biosensors and neuroprosthetics},
  author={Mikhaylov, Alexey and Pimashkin, Alexey and Pigareva, Yana and Gerasimova, Svetlana and Gryaznov, Evgeny and Shchanikov, Sergey and Zuev, Anton and Talanov, Max and Lavrov, Igor and Demin, Vyacheslav and others},
  journal={Frontiers in neuroscience},
  volume={14},
  pages={358},
  year={2020},
  publisher={Frontiers Media SA}
}

@article{khan2025review,
  title={Review of deep learning models with Spiking Neural Networks for modeling and analysis of multimodal neuroimaging data},
  author={Khan, Ayesha and Shim, Vickie and Fernandez, Justin and Kasabov, Nikola K and Wang, Alan},
  journal={Frontiers in Neuroscience},
  volume={19},
  pages={1623497},
  year={2025},
  publisher={Frontiers Media SA}
}

@article{Voytek2015,
  author    = {Voytek, Bradley and Kramer, Mark A. and Case, Jacob and Lepage, Kyle Q. and Tempesta, Zachary R. and Knight, Robert T. and Gazzaley, Adam},
  title     = {Age-Related Changes in 1/f Neural Electrophysiological Noise},
  journal   = {Journal of Neuroscience},
  year      = {2015},
  volume    = {35},
  number    = {38},
  pages     = {13257--13265},
  doi       = {10.1523/JNEUROSCI.2332-14.2015}
}

@article{meghdadi2021resting,
  title={Resting state {EEG} biomarkers of cognitive decline associated with {Alzheimer}’s disease and mild cognitive impairment},
  author={Meghdadi, Amir H and Stevanovi{\'c} Kari{\'c}, Marija and McConnell, Marissa and Rupp, Greg and Richard, Christian and Hamilton, Joanne and Salat, David and Berka, Chris},
  journal={PloS one},
  volume={16},
  number={2},
  pages={e0244180},
  year={2021},
  publisher={Public Library of Science San Francisco, CA USA}
}

@article{dun2022bi,
  title={Bi-directional associations of epilepsy with dementia and {Alzheimer}’s disease: a systematic review and meta-analysis of longitudinal studies},
  author={Dun, Changchang and Zhang, Yaqi and Yin, Jiawei and Su, Binbin and Peng, Xiaobo and Liu, Liegang},
  journal={Age and ageing},
  volume={51},
  number={3},
  pages={afac010},
  year={2022},
  publisher={Oxford University Press}
}

\end{document}